\documentclass{article}
\usepackage{spconf,amsmath,epsfig,bm,amssymb,footmisc, url,subcaption}

\let\OLDthebibliography\thebibliography
\renewcommand\thebibliography[1]{
  \OLDthebibliography{#1}
  \setlength{\parskip}{0pt}
  \setlength{\itemsep}{0pt plus 0.3ex}
}

\begin{document}\sloppy

\def\x{{\mathbf x}}
\def\L{{\cal L}}

\title{CTM: Collaborative Temporal Modeling for Action Recognition}

\name{Qian Liu, Tao Wang, Jie Liu, Yang Guan, Qi Bu, Longfei Yang}

\address{iQIYI, Inc.}

\maketitle

\begin{abstract}
With the rapid development of digital multimedia, video understanding has become an important field. For action recognition, temporal dimension plays an important role, and this is quite different from image recognition. 
In order to learn powerful 
feature of videos, 
we propose a Collaborative Temporal Modeling (CTM) block (Figure \ref{fig1}) to learn temporal information for action recognition.
Besides a parameter-free identity shortcut, as a separate temporal modeling block, CTM includes two collaborative paths: a spatial-aware temporal modeling path, which we propose the Temporal-Channel Convolution Module (TCCM) with unshared parameters for each spatial position ($H \times W$) to build, and a spatial-unaware temporal modeling path. 
CTM blocks can seamlessly be inserted into many popular networks to generate CTM Networks and bring the capability of learning temporal information to 2D CNN backbone networks, which only capture spatial information.
Experiments on several popular action recognition datasets demonstrate that CTM blocks bring the performance improvements on 2D CNN baselines, and our method achieves the competitive results against the state-of-the-art methods.
Code will be made publicly available.

\end{abstract}


\section{Introduction}
\label{sec:intro}

As a result of the rapid development of digital multimedia,
action recognition has become an important research field for video understanding, video surveillance, etc. 
For action recognition, temporal dimension plays an important role, which is different from image recognition, since that the information of a single frame is often not enough to recognize the action class. 
In this paper, we will focus on how to learn the temporal information from videos.

\begin{figure}
	\begin{center}
		\includegraphics[width=0.9\linewidth]{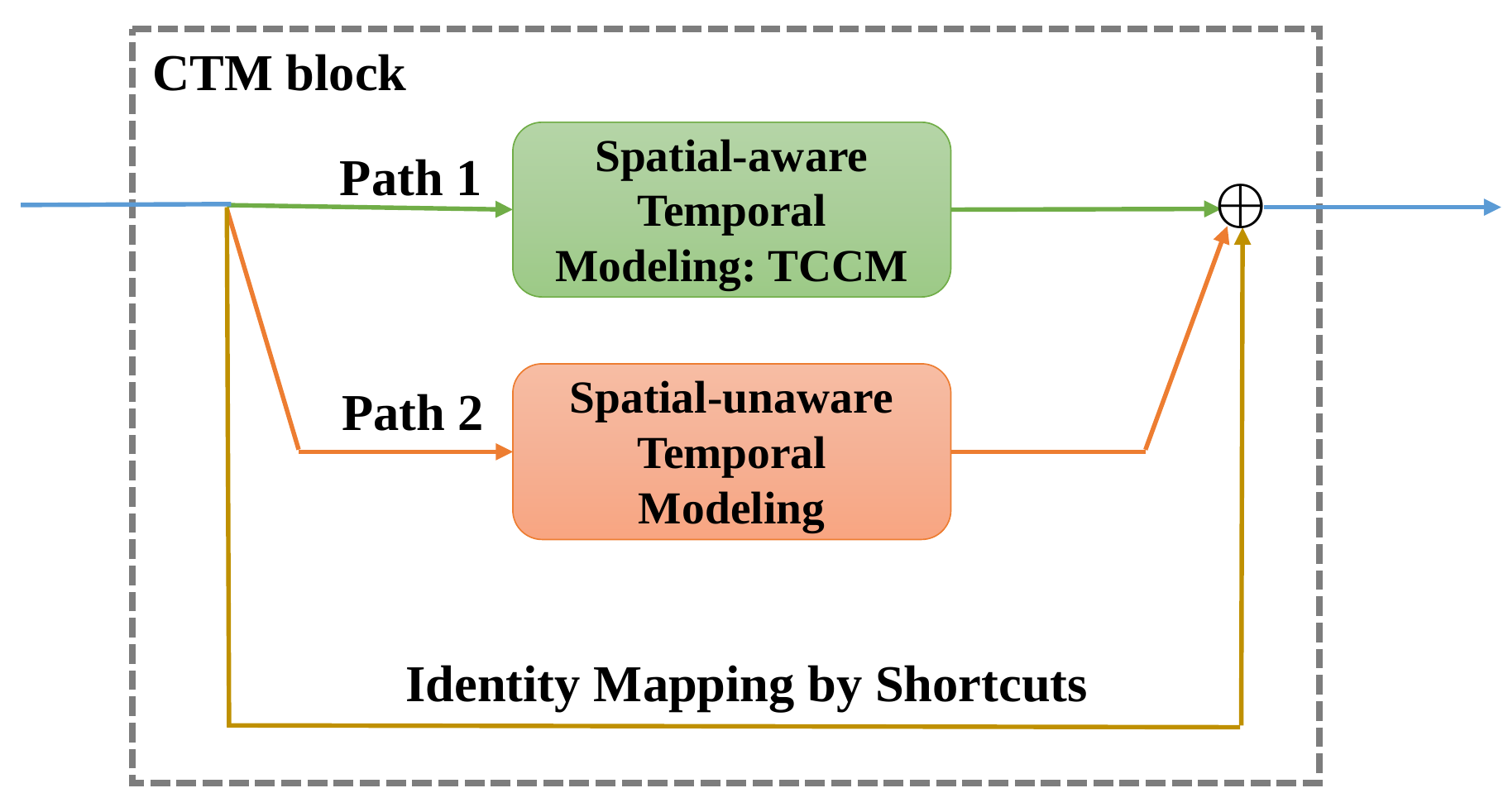}
		\caption{The Collaborative Temporal Modeling (CTM) block. Besides a parameter-free identity shortcut, the CTM block includes two paths : Path 1 is a spatial-aware temporal modeling path: TCCM, and Path 2 is a spatial-unaware temporal modeling path. ``$\bigoplus$" denotes element-wise addition.} \label{fig1}
	\end{center}
\end{figure}

Recently, action recognition has witnessed good improvements resulted from deep convolutional neural networks (CNNs), and the existing deep learning based approaches can be divided into two major categories: 2D CNNs based methods and 3D CNNs based methods. A lot of methods adopt 2D CNNs as the backbones for action recognition. 2D CNNs extract the spatial feature representation of each frame. The late fusion step is used for aggregating frame-wise high-level features to prediction.
It is argued that late fusion ways cannot well learn the temporal information due to the high-level feature is very abstracted. Some late fusion ways (e.g. maximum pooling, average pooling) even do not learn the temporal information.
3D convolution kernel encodes the spatial and temporal features jointly.  
Some methods \cite{qiu2017learning,tran2018closer,xie2018rethinking} decompose 3D convolution kernel into separate spatial and temporal part.
These 3D CNN based methods can learn spatio-temporal representation and achieve superior performance. 

2D CNNs backbones have the advantage of low computational cost but only learn spatial information. 3D CNNs based methods learn spatio-temporal representation but have higher computational burdens than 2D CNNs.
It is still an active field that find proper network architectures to learn the spatial and temporal feature from videos.

Let us focus on the temporal information learning ways. 
In most cases, a block (e.g. a P3D-A block) conducts both the temporal information learning and the spatial information learning. 
Although the spatial modeling and temporal modeling are factorized as different layers in some methods (we name this ``layer decoupling", and a toy example is shown in Figure \ref{fig2}), they are still operated in the same block.
We expect that a separate temporal modeling block may have strong ability of learning temporal feature representation, 
e.g. the spatio-temporal encoding blocks are decomposed into separate spatial modeling blocks and separate temporal modeling blocks (we name this ``block decoupling" and show a toy example in Figure \ref{fig2}).
The intuition here is that different blocks should not have the same modeling capacity, analogous to the human retina's working mechanism that neurons (Parvo cells and Magno cells) in the retina do not have the same capabilities. 

Besides, the parameters of a temporal filter are shared for each spatial position ($H \times W$)  in most cases, thus the temporal filter is spatial-unaware.
However, spatial-aware temporal modeling may be quite effective for action recognition, due to that the motion directions and magnitudes at different positions ($H \times W$) are different for two consecutive frames.  
We consider that the spatial-aware temporal modeling will be collaborative to the existing spatial-unaware temporal modeling.

Based on the above motivations,
we propose the Collaborative Temporal Modeling (CTM) block to learn the temporal information for action recognition. 
It is a separate temporal modeling block. 
As shown in Figure \ref{fig1}, besides a parameter-free identity shortcut, CTM block includes two paths: Path 1 is a spatial-aware temporal modeling path shown in Figure \ref{fig3}, and Path 2 is a spatial-unaware temporal modeling path shown in Figure \ref{fig4}. 
We propose a spatial-aware Temporal-Channel Convolution Module (TCCM) to build Path 1. The parameters of temporal filters in TCCM are unshared for each spatial position ($H \times W$).
In CTM block, spatial-aware Path 1 and spatial-unaware Path 2 are collaborative to learn the temporal feature.
The CTM blocks are flexible and can be inserted into many popular network architectures to generate CTM networks with very limited additional computational cost. 
Specifically, for 2D CNN backbones, which only have spatial modeling ability, CTM blocks bring the high capability of temporal modeling to them. 

To summarize, the main contributions of our paper include:
1) We propose a simple yet effective Collaborative Temporal Modeling (CTM) block to operate separate temporal modeling.
2) We propose a Temporal-Channel Convolution Module (TCCM) to learn spatial-aware temporal encoding.
3) We propose CTM networks with high capability of learning spatial and temporal features for action recognition.
4) Experiments demonstrate that CTM blocks bring performance improvements on 2D CNN baselines and our method achieves the competitive results against the state-of-the-art methods on several action recognition datasets.

\section{Related Work}

Deep learning-based methods \cite{luo2019grouped,wang2016temporal,he2019stnet,lin2019tsm,jiang2019stm,zolfaghari2018eco,Chen_2018_ECCV} have superior performance than hand-crafted features and obtain remarkable success on some public datasets \cite{soomro2012ucf101,kay2017kinetics,Kuehne11HMDB}.
Action recognition methods based on CNNs can be grouped into two categories: 2D CNNs based methods \cite{karpathy2014large,wang2016temporal} and 3D CNNs based methods. 

With the input frames which are selected from a video by some specific regulations, 2D CNNs operate on each frame independently.
Temporal Segment Networks (TSN) \cite{wang2016temporal} is one of the representative 2D CNN methods. 
2D CNNs extract the spatial feature of each frame, then the late fusion step
can aggregate each frame's spatial feature to prediction. 
Some late fusion ways (e.g. maximum pooling, average pooling) cannot learn the temporal information. 
Other late fusion ways specially designed for involving temporal relation often cannot well learn the temporal information. This may result from that the high-level feature is highly abstracted. 

\begin{figure}	
	\centering
	\begin{subfigure}[t]{0.45\linewidth}
		\centering
		\includegraphics[width=\linewidth,height=14ex]{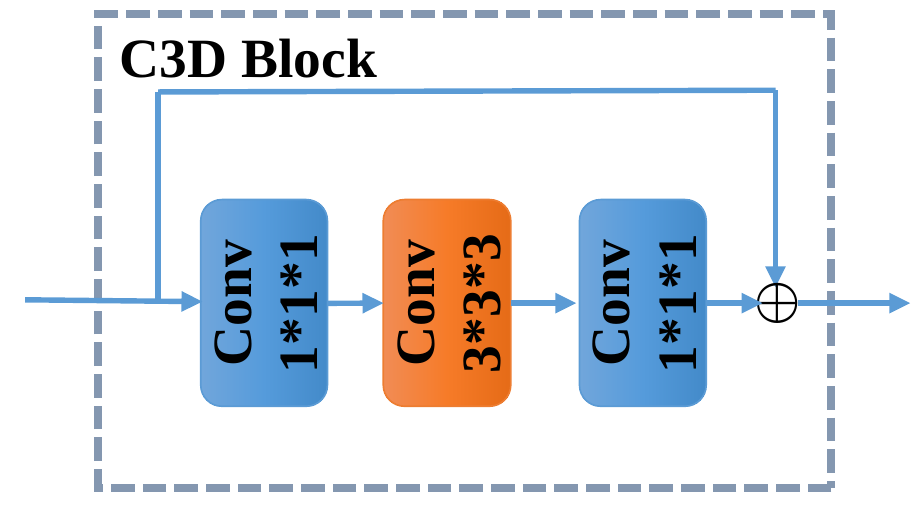}
		\caption{}\label{fig:1a}		
	\end{subfigure}
	\begin{subfigure}[t]{0.45\linewidth}
		\centering
		\includegraphics[width=1.1\linewidth,height=15ex]{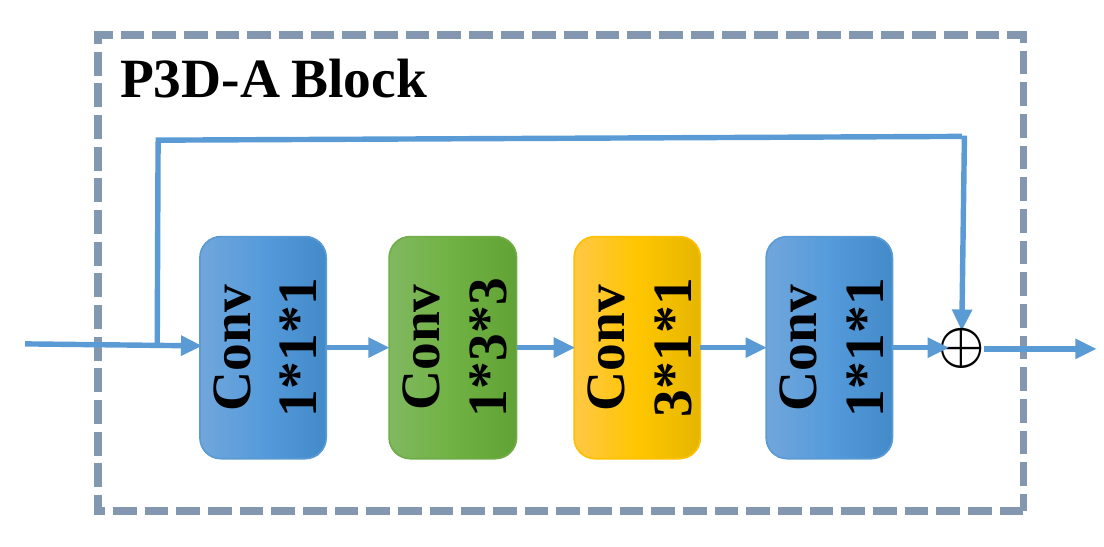}
		\caption{}\label{fig:1b}
	\end{subfigure}
	\begin{subfigure}[t]{0.45\linewidth}
	\centering
	\includegraphics[width=\linewidth]{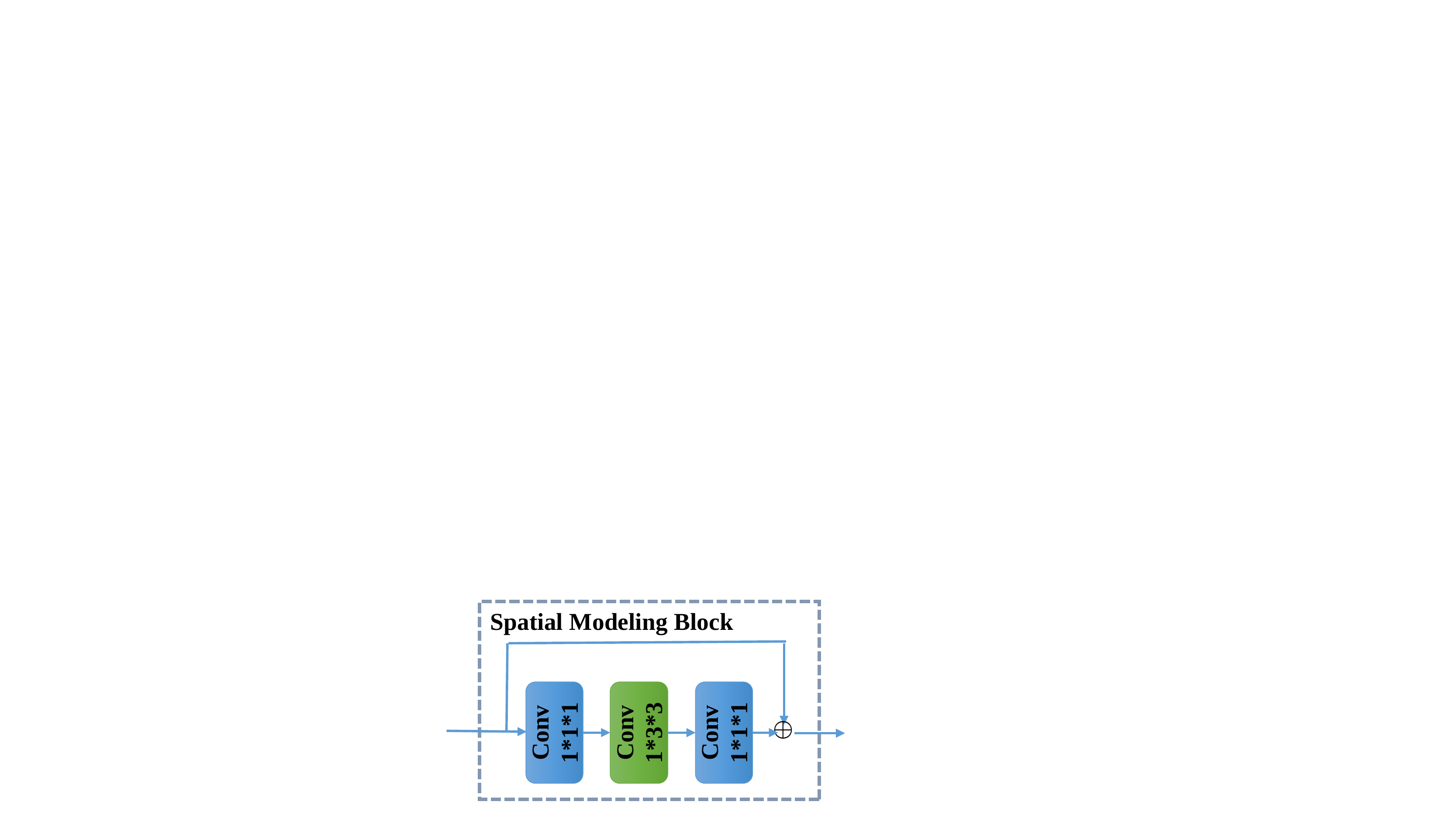}
	\caption{}\label{fig:1c}
\end{subfigure}
	\begin{subfigure}[t]{0.45\linewidth}
	\centering
	\includegraphics[width=\linewidth]{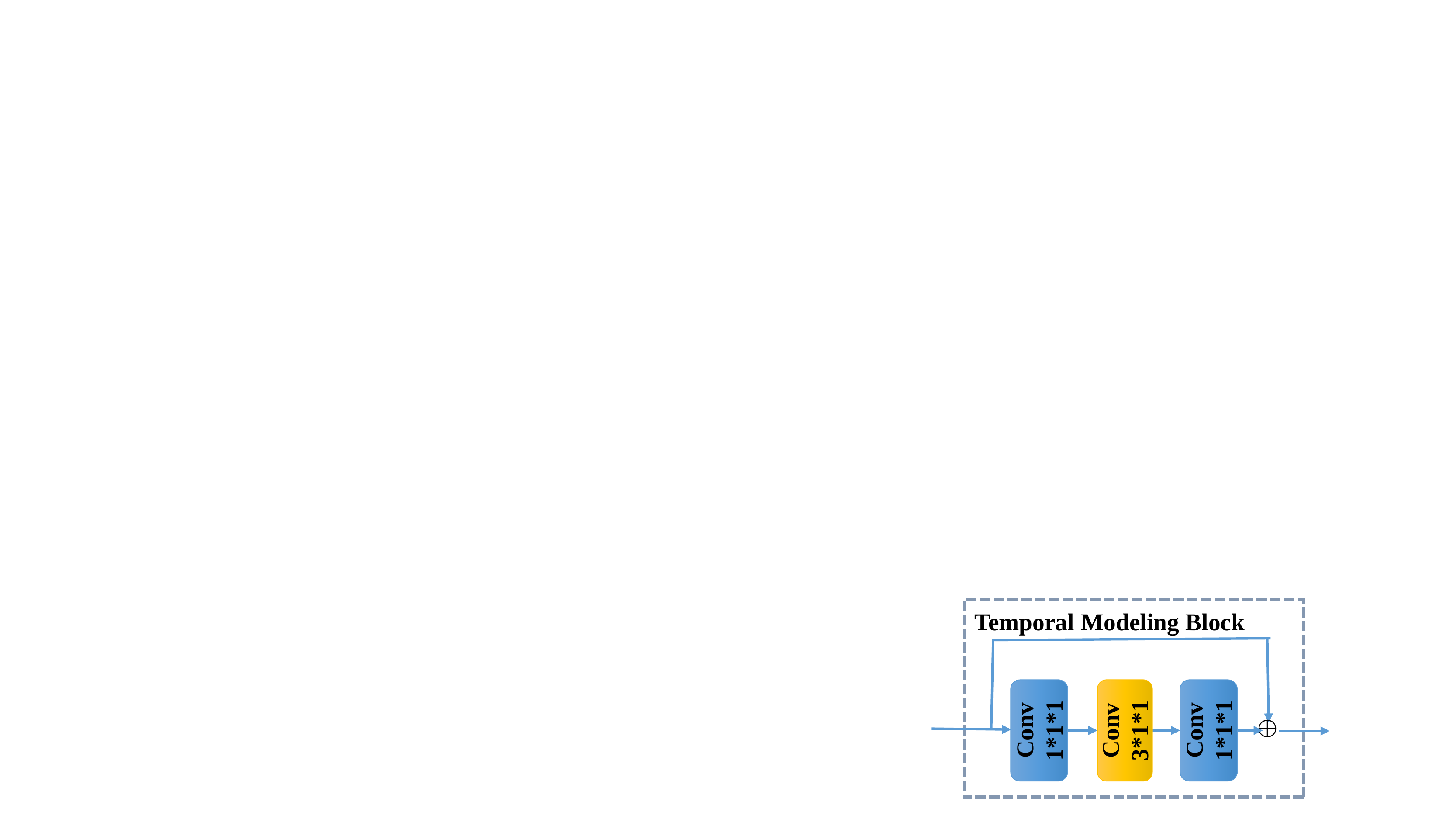}
	\caption{}\label{fig:1d}
\end{subfigure}
	\caption{Toy examples of layer decoupling and block decoupling. (a) a C3D block. (b) a P3D block. P3D decouples the 3D convolution layer into the spatial convolution layer and the temporal convolution layer. (c) a separate spatial modeling block. (d) a separate temporal modeling block. From (a) to (b) is a toy example of layer decoupling. From (b) to ((c) + (d)) is a toy example of block decoupling.}\label{fig2}
\end{figure}

3D convolution networks can effectively extract spatial-temporal information.
C3D \cite{tran2014c3d} encodes the spatial and temporal features jointly via a 3D convolution kernel $D \times K \times K$, where $D$ is the temporal dimension and $K$ is the spatial size. I3D \cite{carreira2017quo} 
inflates parameters of pretrained 2D image classification models to 3D.
The Channel-Separated Convolutional Network (CSN) \cite{tran2019video} replaces expensive 3D convolution operations by channel interactions and spatio-temporal interactions separately. 
Considering the high computational burdens of 3D convolution kernel, in some methods, such as Pseudo-3D \cite{qiu2017learning}, R(2+1)D \cite{tran2018closer} and S3D \cite{xie2018rethinking}, a 3D convolution kernel $D \times K \times K$ is decoupled into a spatial convolution kernel $1 \times K \times K$ and a temporal convolution kernel $D \times 1 \times 1$,
and simulate a 3D convolution layer is decoupled to a spatial encoding layer and a temporal encoding layer. 
However, after such ``layer decoupling" (a toy example is shown in Figure \ref{fig2}), the spatial encoding layer and the temporal encoding layer are still operated in the same block. 
Besides, the parameters of temporal filters are not spatial-aware in most methods.

\section{Approach}

In this section, firstly Temporal-Channel Convolution Module (TCCM) in CTM is described. Then, CTM blocks and CTM networks are introduced.

\begin{figure}
	\begin{center}
		\includegraphics[width=0.95\linewidth]{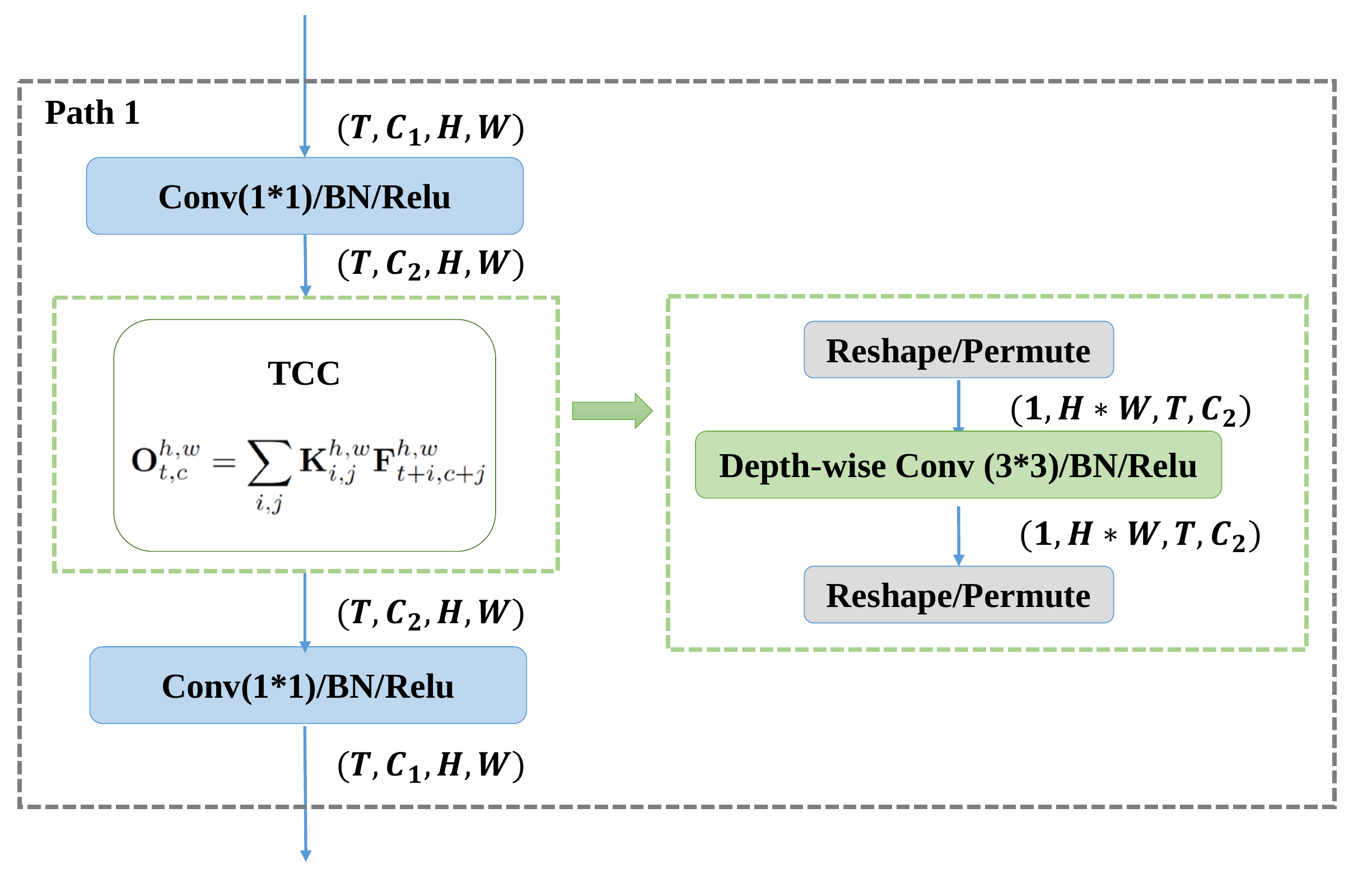}
		\caption{Temporal-Channel Convolution Module (TCCM). TCCM is ``Path 1" of a CTM block. For feature maps $ {\bf F} $ with the size of $T \times C\times H \times W$, TCC operates on the $T$ and $C$ dimensions.} \label{fig3}
	\end{center}
\end{figure}

\subsection{Temporal-Channel Convolution Module (TCCM)}\label{3-1-tccm}

At first, we describe the design of the Temporal-Channel Convolution (TCC) in TCCM. Consider the input feature maps with the size of $T \times C \times H \times W$ (we omit batch-size for simple), where $T, C, H, W$ are the temporal length, 
 channels, height and width, respectively, TCC does not operate on the $H$ and $W$ dimensions. Instead, TCC operates on the $T$ and $C$ dimensions. 
TCC is designed for temporal modeling with unshared parameters for each spatial position ($H\times W$).
Formally, the Temporal-Channel Convolution (TCC) can be formulated as:
\begin{equation}{\bf O}^{h,w}_{t,c} = \sum_{i,j} {\bf K}^{h,w}_{i,j} {\bf F}^{h,w}_{t+i,c+j},\end{equation}\label{eq1} 
where $ {\bf K}^{h,w}_{i,j} $ denotes the filter for a spatial position ($h \times w$),
the subscripts $i, j$ denote the temporal and channel indices of the kernel,
$ {\bf F} $ is the input feature maps,
and ${\bf O}$ is the output feature maps.
Eq \ref{eq1} can be implemented easily via combining some existing functions in some popular deep learning platforms. Specifically, the shape of feature maps is changed from $T \times C \times H \times W$ to $1 \times HW \times T \times C$ and the common depth-wise 2D convolution is operated on this feature maps, then the shape is changed back.
The architecture of TCCM is shown in Figure \ref{fig3}. Firstly, in order to decrease the computational cost, the 1x1 convolution layer is leveraged for reduce the number of channels $C_{1}$ of the input feature maps to $C_{2}$. Then, TCC is operated. Finally, restore the number of channels from $C_{2}$ to $C_{1}$.             
In this paper, the 2D TCC kernel size is set to $3 \times 3$, and the shape of kernel is $(H \times W,1,3,3)$.

\subsection{Collaborative Temporal Modeling (CTM) block}

\begin{figure}
	\begin{center}
		\includegraphics[width=0.45\linewidth]{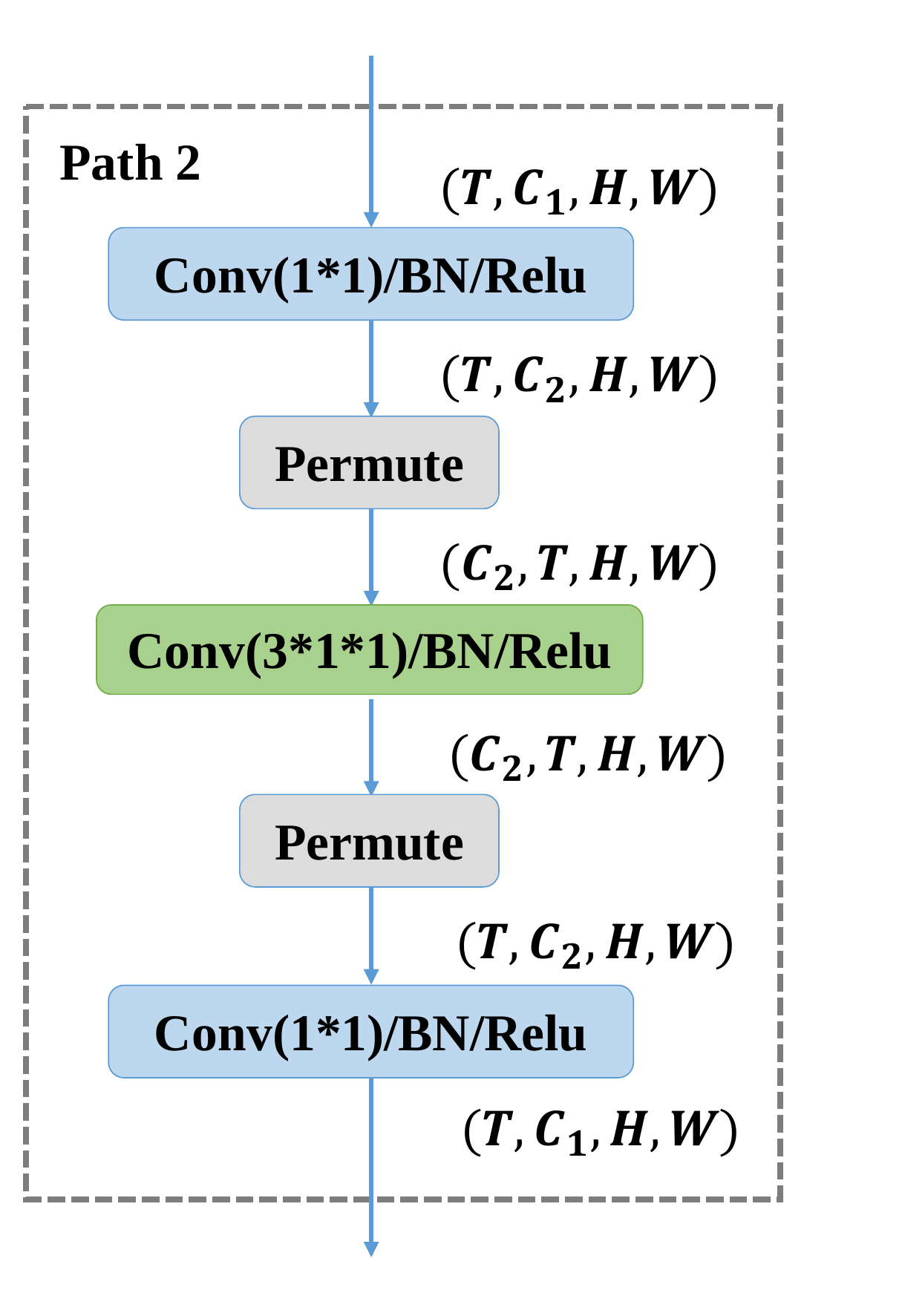}
		\caption{The architecture of ``Path 2" in a CTM block.} \label{fig4}
	\end{center}
\end{figure}

Figure \ref{fig1} presents the architecture of the Collaborative Temporal Modeling (CTM) block.
It is a separate temporal modeling block. Besides a parameter-free identity shortcut, CTM includes two collaborative temporal modeling paths.  
The TCCM path (Path 1) is spatial-aware temporal modeling module and is described in Section \ref{3-1-tccm}.
Another temporal encoding path (Path 2) is a module with the $3\times1\times1$ temporal convolution and this module is a spatial-unaware temporal modeling module, as illustrated in Figure \ref{fig4}. 
In Figure \ref{fig4}, after reducing the number of channels $C_{1}$ of the input feature maps to $C_{2}$, the shape of feature maps is permuted to $C_{2} \times T \times H \times W$ to match the $3\times1\times1$ temporal convolution. Then, the shape is permuted back and the number of channels is restored from $C_{2}$ to $C_{1}$ after this $3\times1\times1$ temporal convolution layer. In Path 2, the parameters of $3\times1\times1$ temporal convolution are shared for each spatial position ($H \times W$).
Path 1, Path 2 and a parameter-free identity shortcut are combined to build a CTM block.

\begin{figure}
	\begin{center}
		\includegraphics[width=\linewidth]{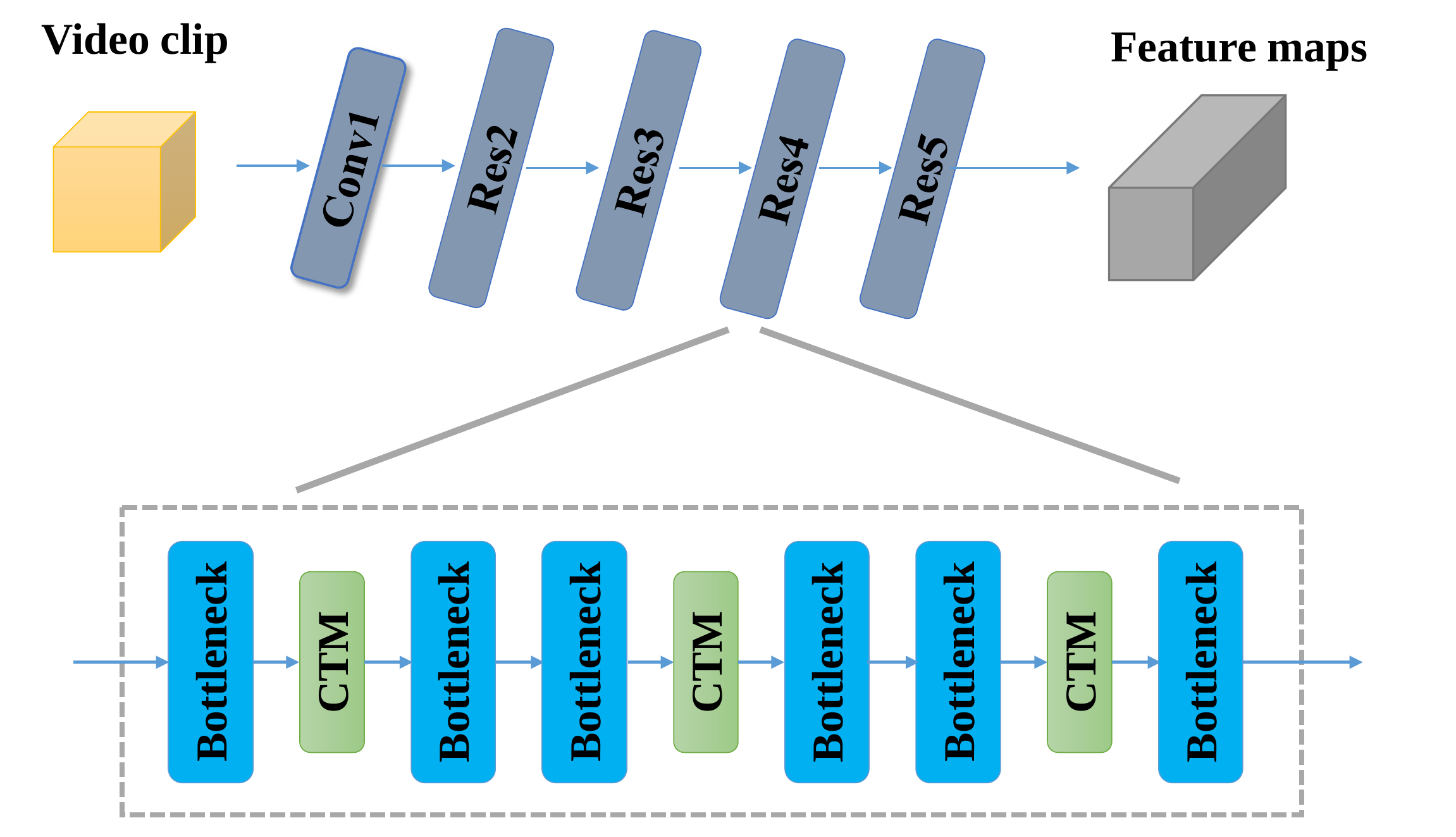}
		\caption{The overall architecture of CTM network. 2D ResNet-50 backbone is adopted as an example. Three CTM blocks (after the 1st, 3rd and 5th bottleneck block, respectively) are added to Res4. ``Bottleneck" denotes a residual block of 2D ResNet-50 backbone and it is a separate spatial modeling block. ``CTM" denotes a CTM block and it is a separate temporal modeling block. ``Bottleneck" blocks and ``CTM" blocks are combined in CTM networks to make CTM networks have high capability of spatial and temporal modeling.} \label{fig5}
	\end{center}
\end{figure}

\subsection{CTM Network}

\begin{table*}
	\begin{center}
	\end{center}
\end{table*}

The CTM block can seamlessly be injected into many popular networks to generate CTM networks. CTM blocks can bring the capability of learning temporal information to 2D CNN backbone networks.
Here, we adopt 2D ResNet-50 as an example, and the overall architecture of a CTM network with 2D ResNet-50 backbone is presented in Figure \ref{fig5}. We add three CTM blocks (after the 1st, 3rd and 5th bottleneck block, respectively) to Res4 in the 2D ResNet-50 backbone.

\section{Experiment}

\subsection{Datasets}
Here, we evaluate the performance of CTM on the several popular datasets: Mini-Kinetics-200 \cite{xie2018rethinking}, Kinetics-600 \cite{carreira2018short}, UCF-101 \cite{soomro2012ucf101} and HMDB51 \cite{Kuehne11HMDB}.

Mini-Kinetics-200 used in \cite{xie2018rethinking} is a subset of Kinetics-400 \cite{kay2017kinetics}, which is a large-scale and popular action recognition benchmark. In Mini-Kinetics-200, 200 action classes are contained.
Mini-Kinetics-200 has 80000 videos in the train set and 5000 videos in the validation set individually. 
\footnote {The training/validation split files are provided on \url{https://github.com/s9xie/Mini-Kinetics-200}.}
Thus, we can train and test models faster and more easily on Mini-Kinetics-200, while it still provides the enough data diversity for the models training. 
Kinetics-600 \cite{carreira2018short} is a larger scale and newly released dataset and it consists of 480k videos from 600 action categories.
Kinetics-600 has 390k training examples and 30k validating examples.
\footnote{We actually only downloaded about 375k videos in the train set and about 28700 videos in the validation set in Kinetics-600 due to that some videos' youtube\_ids are no longer available. For Mini-Kinetics-200, we actually have only about 74000 videos in the train set and about 4700 videos in the validation set for the same reason.}
UCF-101 \cite{soomro2012ucf101} includes 13320 videos divided into 101 action categories,
and HMDB51 \cite{Kuehne11HMDB} contains around 7000 videos distributed from 51 classes.

\subsection{Implementation Details}

\begin{table}
	\begin{center}
		\caption{Performance comparisons on Mini-Kinetics-200. We add 1 block to Res4 into 2D ResNet-50 backbone.}
		\label{tab1}
		\begin{tabular}{|l|c|}
			\hline
			Model, 2D Resnet50 &Val Acc top-1 (\%)\\
			\hline
			baseline  & 76.76 \\
			TCCM + shortcut (Path1 + shortcut) & 78.60 \\ 
			Path 2 + shortcut & 78.58  \\ 
			CTM (Path1 + Path 2 + shortcut)& \bf 78.94 \\
			\hline
		\end{tabular}
	\end{center}
\end{table}

\begin{table}
	\begin{center}
		\caption{The results of adding the CTM block into different stages. One CTM block is added into different stages (Res2, Res3, Res4 and Res5) in 2D ResNet-50 backbone}
		\label{tab2}
		\begin{tabular}{|l|c|}
			\hline
			Model, 2D Resnet50 &Val Acc top-1 (\%)\\
			\hline
			baseline  & 76.76 \\
			Res2  & 78.87 \\
			Res3 & 78.79 \\
			\bf Res4 & \bf 78.94 \\
			Res5 & 78.07 \\
			\hline
		\end{tabular}
	\end{center}
\end{table}

{\bf Training}
In the training stage, the input of our model is a 8-frame input clip, which is sampled by a fix frame interval from 64 consecutive frames randomly clipped from the original full-length video.
We use SGD algorithm with a momentum of 0.9 and a weight decay of 0.0001.
For Mini-Kinetics-200, we train our model for 60 epochs and the batch size is 96 on 4 GPUs. The initial learning rate is 0.02 and decays by 0.1 at epoch 30 and epoch 50. 
Since Kinetics-600 is a larger dataset, the model is trained for 100 epochs. The initial learning rate is 0.02 and decays by 0.1 at epoch 50 and epoch 80. The batch-size is 128 on 8 GPUs.
For UCF-101 and HMDB51, we adopt Kinetics-600 pre-trained models. 
We use an initial learning rate of 0.002, and the learning rate is decayed by a factor 10 every 10 epochs. The models are trained for 30 epochs. All the BN layers are fixed except the first one. 
 
In the rest of this paper, unless specified, our backbones are pretrained on ImageNet. 
In order to maintaining initial behavior of backbones at the beginning of training, parameters of the last BN layers in Path 1 and Path 2 are initialized as zero to make the whole CTM block an identity mapping.

{\bf Inference}
We largely follow the inference protocol in \cite{wang2018non}. Top-1 accuracy on the validation set will be reported. The spatial size with $224 \times 224$ pixels is used for inference. For UCF-101 and HMDB51, we 
report the average accuracy across three splits.

\begin{table}
	\begin{center}
		\caption{The results of adding 1, 3, and 6 CTM blocks into Res4 of 2D ResNet-50 backbone.}
		\label{tab3}
		\begin{tabular}{|l|c|}
			\hline
			Model, 2D Resnet50 &Val Acc top-1 (\%)\\
			\hline
			baseline  & 76.76 \\
			+ 1 CTM block & 78.94 \\
			+ 3 CTM blocks & 79.13 \\
			\bf + 6 CTM blocks & \bf  79.15 \\
			\hline
		\end{tabular}
	\end{center}
\end{table}

\begin{table}
	\begin{center}
		\caption{The results on Kinetics-600. (Val Acc top-1 (\%))}
		\label{tab4}
		\begin{tabular}{|l|c|c|c|}
			\hline
			Models &input  & Acc \\
			\hline
			TSN \cite{wang2016temporal} (SE-ResNeXt152) & RGB  & 76.16 \\ 
			StNet (SE-ResNeXt101) \cite{he2019stnet}   & RGB &   76.04 \\
			StNet (IRv2) \cite{he2019stnet} & RGB    & \bf 78.99 \\
			LGD-2D \cite{qiu2019learning} &  RGB  &76.7 \\
			LGD-3D \cite{qiu2019learning}& RGB  & 78.3\\
			LGD-3D (Long) \cite{qiu2019learning} & RGB  &\bf 81.5\\ 
			D3D \cite{stroud2018d3d}& RGB &77.9 \\
			D3D+S3D-G \cite{stroud2018d3d}& RGB & 79.1 \\
			Oct-I3D \cite{chen2019drop} & RGB  & 76.0 \\
			\bf (Our) CTM & \bf RGB  & \bf 78.24 \\
			\hline			
		\end{tabular}
	\end{center}
\end{table}

\subsection{An Ablation Study of CTM Networks}
Here, we evaluate the performance of CTM on Mini-Kinetics-200 with the backbone of 2D ResNet50.
In Table \ref{tab1}, we add a single block before the last residual block of the Res4 stage. 
In Table \ref{tab1}, the original 2D Resnet50 baseline achieves $76.76\%$ accuracy.
TCCM+shortcut exhibits better performance with the accuracy of $78.60\%$. 
CTM obtain the best accuracy of $78.94\%$ and further improve 2D CNN Baseline's accuracy by about $2.18\%$.  
This demonstrates that the action recognition result is significantly benefit from CTM blocks.

\paragraph*{Which stage to add the CTM block?}

Here, a single CTM block is added into different stages (Res2, Res3, Res4 and Res5) in 2D ResNet-50 backbone, respectively.  
Table \ref{tab2} shows the results, and the accuracies are $78.87\%$, $78.79\%$, $78.94\%$ and $78.07\%$, respectively. 
Note that all of the results are better than baseline, and    
the improvement on Res2, Res3, and Res4 is similar. 
For Res5, the top-1 accuracy declines, and this may be resulted from the high-level feature map in Res5 is highly abstract.
In the rest of this paper, we add CTM blocks into Res4 by default. 

\paragraph*{How many CTM blocks?}

We add one CTM block (before the last residual block), three CTM blocks (after the 1st, 3rd and 5th residual block, respectively) and six CTM blocks (after every residual block) to Res4 in 2D ResNet-50 backbone, respectively.
In Table \ref{tab3}, CTM Network with one CTM block achieves $78.94\%$ top-1 accuracy. 
CTM Networks with three CTM blocks and five CTM blocks obtain the top-1 accuracy $79.13\%$ and $79.15\%$, respectively.
We use CTM Network with three CTM blocks in the following experiments since they provide good performance and calculate fast.

\begin{table}
	\begin{center}
		\caption{The results on UCF-101 and HMDB51. (Val Acc top-1 (\%))} 
		\label{tab5}
		\begin{tabular}{|l|c|c|c|}
			\hline
			Models &input & UCF-101 & HMDB51\\
			\hline
			StNet \cite{he2019stnet}& RGB &95.7 & -\\ 
			STM \cite{jiang2019stm} & RGB  & 96.2 &72.2\\ 
			TSM \cite{lin2019tsm}& RGB  & 96.0  &73.6\\  
			I3D \cite{carreira2017quo} & RGB & 95.6 & 74.8\\
			R(2+1)D \cite{tran2018closer} & RGB & 96.8  &74.5\\
			S3D-G \cite{xie2018rethinking} & RGB & 96.8 & 75.9\\  
			\bf (Our) CTM & RGB & \bf 96.9& \bf 74.6\\                
			\hline
		\end{tabular}
	\end{center}
\end{table}

\begin{table*}
	\begin{center}
		\caption{Our method consistently outperforms 2D baselines on multiple datasets.}
		\label{tab-Improv}
		\begin{tabular}{|l|c|c|c|c|}
			\hline
			Datasets & Kinetics-600  & UCF-101 & HMDB51 &Mini-Kinetics-200\\			
			(Backbone)& (2D ResNet-101) & (2D ResNet-101) & (2D ResNet-101)& (2D ResNet-50)\\ 
			\hline
			baseline & 75.34  & 95.5 & 69.0  & 76.76 \\ 
			\bf (Our) CTM & \bf 78.24  &\bf 96.9  &\bf 74.6  &\bf  79.13  \\			
			\hline
		\end{tabular}
	\end{center}
\end{table*}

\subsection{Improvements on 2D CNN Baselines}
We show the improvements on 2D CNN baselines brought by CTM blocks here.  
For 2D ResNet-101 backbone, we add three CTM blocks (after the 6th, 12th and 18th residual block, respectively) to Res4.
For UCF-101 and HMDB51, the Kinetics-600 pretrained model is used.
Table \ref{tab-Improv} details the accuracy improvement on different datasets. 
The results of our methods consistently outperform the 2D CNN baseline. 
It demonstrates that 2D CNN baseline significantly benefit from our CTM blocks.

\subsection{Comparison with State-of-the-Arts}

\paragraph*{Experiments on Kinetics 600}
Table \ref{tab4} summarizes the performance of state-of-the-art methods. CTM network with 2D ResNet-101 backbone obtain the $78.24\%$ top-1 accuracy.
The results demonstrate that CTM network achieves better results than most 2D methods and obtains competitive results against 3D methods. 

\paragraph*{Experiments on UCF-101 and HMDB51}

In Table \ref{tab5}, some state-of-the-art techniques are compared on UCF-101 and HMDB51.
We report the averaged results across three splits. 
In table \ref{tab5}, we can see that CTM networks achieve better results than 2D methods and competitive performance compared with 3D methods.
The top-1 accuracy of our method with only RGB input achieves 96.9\% on UCF-101 and 74.6\% on HMDB51. 

\section{Conclusion}
In this paper, we proposed the Collaborative Temporal Modeling (CTM) block for action recognition. 
In CTM block, we 
proposed Temporal-Channel Convolution Module (TCCM) to build the spatial-aware temporal modeling path.
Furthermore, CTM blocks can be inserted into many existing network architectures to generate CTM networks, and CTM blocks bring the capability of learning temporal information to 2D CNN networks.
Experiments on several popular action recognition datasets demonstrate that CTM blocks bring the performance improvements on 2D CNN networks.

\bibliographystyle{IEEEbib}
\bibliography{myref-mmm}

\end{document}